\documentclass{article}

\usepackage[preprint]{neurips_2026}

\usepackage{times}
\usepackage{latexsym}
\usepackage{subcaption}
\usepackage{microtype}
\usepackage{adjustbox}
\usepackage{multirow}
\usepackage{paralist}
\usepackage[inline]{enumitem}
\usepackage{url}
\usepackage{textcomp}
\usepackage{amssymb}
\usepackage{caption}
\usepackage{color,soul}
\usepackage{paralist}
\usepackage{float}
\usepackage{placeins}
\usepackage{aliascnt}
\usepackage{mathtools}
\usepackage{amsmath}
\usepackage{fontawesome5}
\usepackage{framed}
\usepackage{booktabs}
\usepackage{graphicx}
\usepackage[hidelinks]{hyperref}
\usepackage{algpseudocode}
\usepackage[ruled,vlined]{algorithm2e}
\usepackage[dvipsnames]{xcolor}
\usepackage[T1]{fontenc}
\title{\includegraphics[scale=0.02]{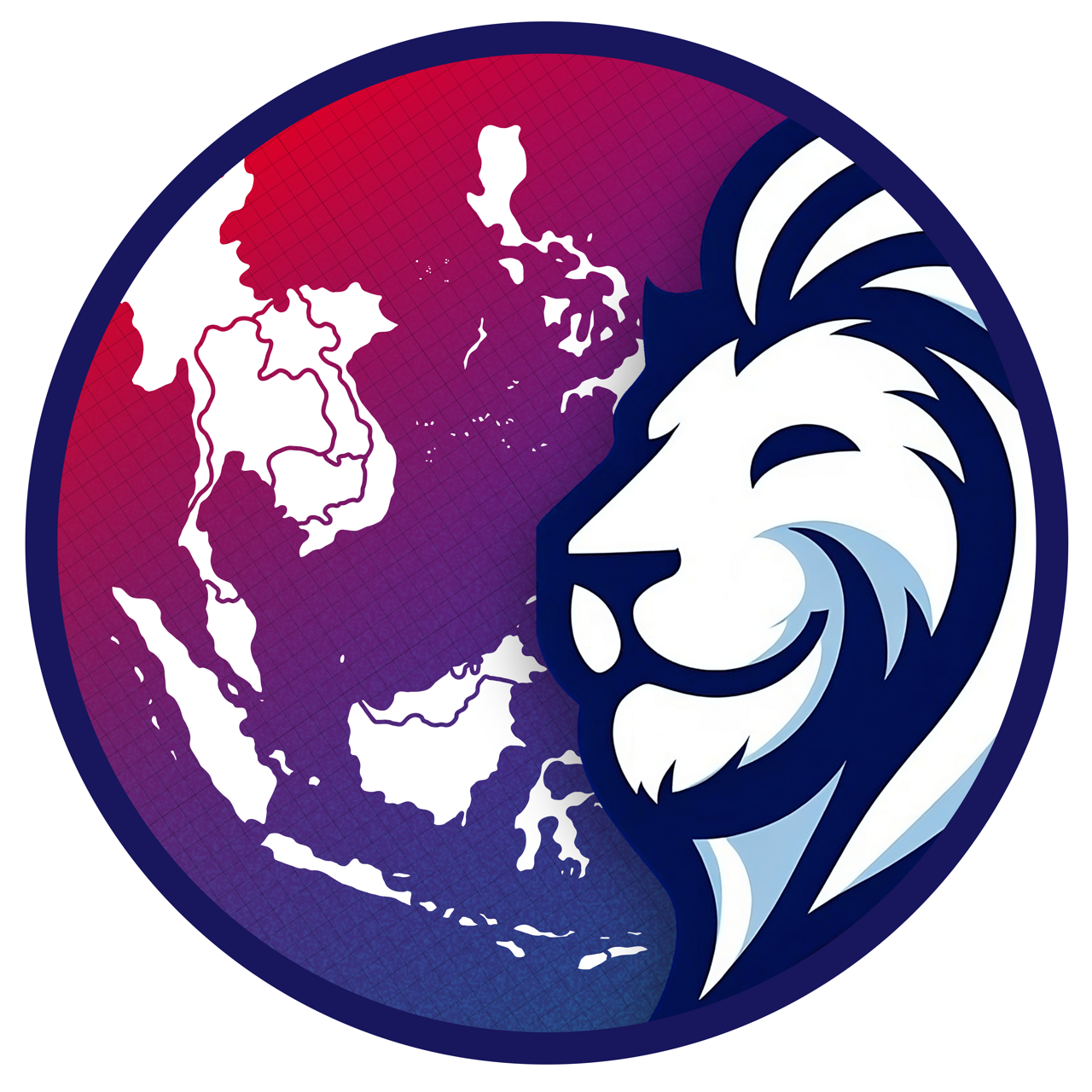}SEA-LION-v4.8: A Technical Report}

\author{%
  AI Products Pillar, AI Singapore \\
  \texttt{sealion@aisingapore.org} \\
\href{https://huggingface.co/collections/aisingapore/sea-lion-v48}{\includegraphics[height=1em]{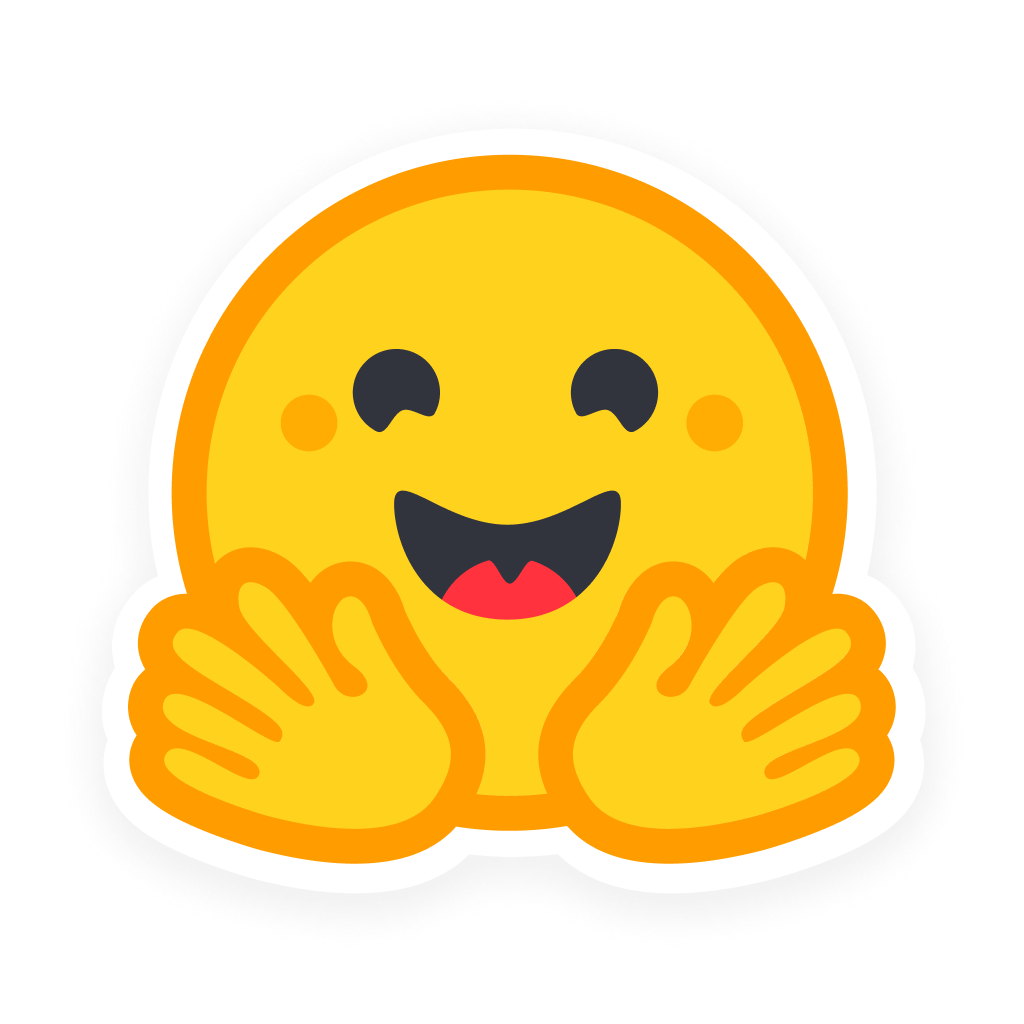}\ SEA-LION-v4.8 Collection}
}

\begin{document}

\maketitle

\begin{abstract}
We introduce \textsc{Nemotron-SEA-LION-v4.8}, a family of Southeast Asian Languages In One Network (SEA-LION) models built upon NVIDIA Nemotron 3.
The family includes 30B-A3B and 120B-A12B models, with both continued-pretrained base checkpoints and post-trained variants.
We adapt the models using Southeast Asian, reasoning, code, and multilingual parallel data, followed by post-training with supervised fine-tuning and online on-policy distillation.
On SEA-HELM, the 30B-A3B model improves the overall SEA score from 46.06 to 51.57, while the 120B-A12B model improves from 49.30 to 63.44.
Across seven Southeast Asian languages, we observe broad capability gains with the 120B-A12B model showing broader and more consistent improvements across tasks.
\end{abstract}

\tableofcontents
\newpage

\section{Introduction}

Large language models (LLMs) have made rapid progress in reasoning, instruction following, coding, long-context understanding, and agentic capabilities. Recent open-weight model families have also become increasingly multilingual, enabling a broader range of users to access capable language models without relying exclusively on proprietary systems. Nevertheless, multilingual coverage remains highly uneven. Languages with substantially less representation in large-scale training corpora often receive weaker linguistic, cultural, and domain-specific support, and improvements observed on English-centric benchmarks do not necessarily transfer uniformly to these languages.

This challenge is particularly important for Southeast Asia, one of the world's most linguistically and culturally diverse regions. The region spans multiple language families and writing systems, with substantial differences in the availability and quality of digital resources. Consequently, even strong multilingual foundation models can exhibit large differences in performance across Southeast Asian languages. Therefore, improving regional capability requires more than simply scaling model size: it requires deliberate choices about language coverage, training data, and training methodology.

In this report, we introduce \textsc{Nemotron-SEA-LION-v4.8}, a new generation of SEA-LION models built upon the NVIDIA Nemotron 3 series. SEA-LION-v4.8 comprises two model scales: a 30B-parameter model with 3B active parameters and a 120B-parameter model with 12B active parameters. For each size, we release both a continued-pretrained base model and a corresponding post-trained model, resulting in four released checkpoints: \textsc{Nemotron-SEA-LION-v4.8-30B-A3B-Base}, \textsc{Nemotron-SEA-LION-v4.8-30B-A3B}, \textsc{Nemotron-SEA-LION-v4.8-120B-A12B-Base}, and \textsc{Nemotron-SEA-LION-v4.8-120B-A12B}.

For the continued-pretraining stage, we adapt the corresponding Nemotron 3 foundation models using 150B high-quality tokens covering question answering, reasoning, code, and multilingual translation data, with particular emphasis on English and Southeast Asian languages. The 30B model is trained using \href{https://github.com/NVIDIA-NeMo/Megatron-Bridge}{Megatron Bridge}, while the 120B model is trained using \href{https://github.com/NVIDIA-NeMo/Automodel}{NVIDIA NeMo AutoModel}. We retain the original Nemotron tokenizers for both model families.
Building on these continued-pretrained checkpoints, we further perform large-scale post-training using our unified training pipeline which consists of online on-policy distillation (OPD) with SFT. Rather than relying exclusively on a static offline instruction dataset, our post-training system allows the model to continuously generate interactions across heterogeneous task environments and learn from supervision provided on its own current behavior. This framework enables the model to combine conventional supervised data with online-generated trajectories while exposing a single evolving model to large numbers of concurrent interaction environments.

\begin{figure}[h!]
    \centering

    \begin{subfigure}[t]{0.48\linewidth}
        \centering
        \includegraphics[width=\linewidth]{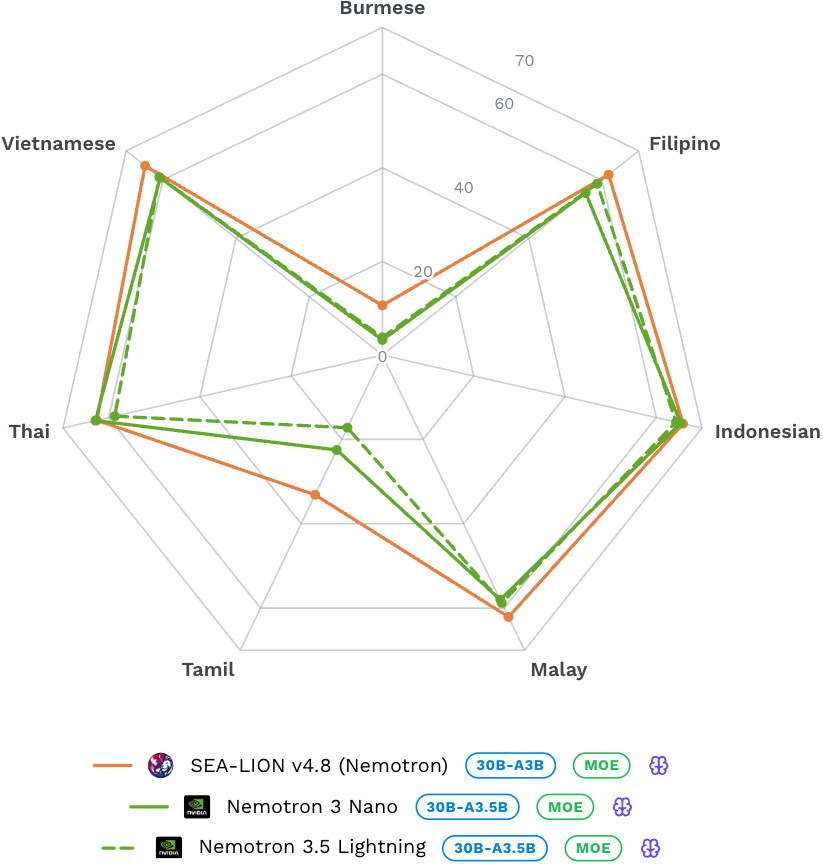}
        \caption{
Overall result comparison between Nemotron-SEA-LION-v4.8-30B-A3B and the two 30B Nemotron baselines: NVIDIA-Nemotron-3-Nano-30B-A3B-BF16 and NVIDIA-Nemotron-3.5-Lightning-30B-A3B-BF16.}
        \label{fig:30b_results}
    \end{subfigure}
    \hfill
    \begin{subfigure}[t]{0.48\linewidth}
        \centering
        \includegraphics[width=\linewidth]{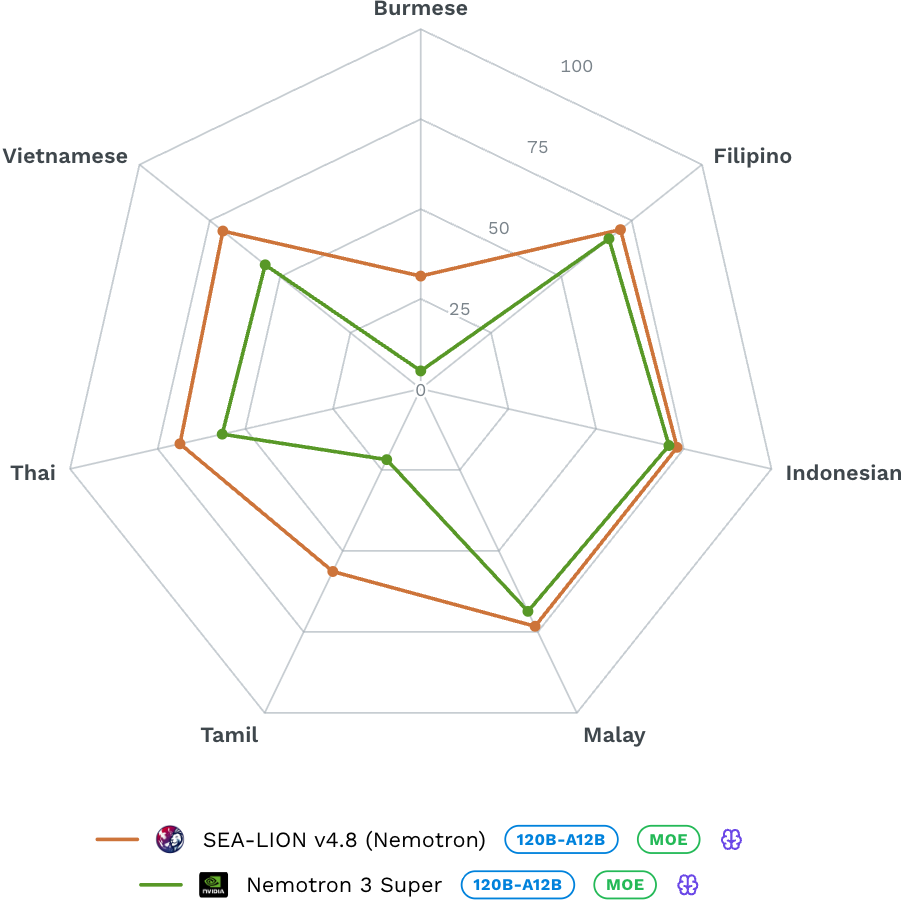}
        \caption{
        The overall result comparison between Nemotron-SEA-LION-v4.8-120B-A12B and its base model (nvidia/NVIDIA-Nemotron-3-Super-120B-A12B-FP8).
        }
        \label{fig:120b_results}
    \end{subfigure}

    \caption{
    Overall result comparisons between Nemotron-SEA-LION-v4.8 models and their respective base models.
    }
    \label{fig:nemotron_results}
\end{figure}

To evaluate the robustness of \textsc{SEA-LION-v4.8}, we use an updated version of SEA-HELM with expanded dataset coverage and improved evaluation methodology.
As shown in Figure~\ref{fig:nemotron_results}, \textsc{SEA-LION-v4.8} consistently improves over the corresponding Nemotron 3 models across Southeast Asian languages while maintaining strong performance across a broad range of capabilities.
In particular, \textsc{Nemotron-SEA-LION-v4.8-30B-A3B} (Figure~\ref{fig:30b_results}) improves the overall SEA-HELM score from 46.06 to 51.57, demonstrating that the proposed continued-pretraining and post-training pipeline is effective at the 30B scale.
At the capability level, the model shows broad improvements in instruction following, natural language generation, reasoning, and language understanding across seven Southeast Asian languages.
The larger \textsc{Nemotron-SEA-LION-v4.8-120B-A12B} model (Figure~\ref{fig:120b_results}) further improves the overall SEA-HELM score from 49.30 to 63.44, with especially large gains in Tamil and Burmese and broader improvements across cultural, instruction-following, knowledge, reasoning, and language-understanding capabilities.
These results suggest that targeted regional adaptation can substantially improve Southeast Asian language capabilities and remains effective when scaling from 30B-A3B to 120B-A12B, without requiring the development of a new foundation model from scratch.

Our contributions are summarized as follows:
\begin{itemize}
    \item We introduce \textsc{Nemotron-SEA-LION-v4.8}, a family of four open models spanning 30B-A3B and 120B-A12B scales, including both pre-trained and instruction-trained checkpoints.
    
    \item We perform continued pre-training on 150B high-quality tokens spanning Southeast Asian and English languages, adapting strong Nemotron 3 foundation models toward regional linguistic and cultural capabilities.
    
    \item We develop a large-scale post-training framework based on online on-policy distillation, enabling a single evolving model to learn continuously from interactions generated across heterogeneous task environments.
\end{itemize}

\begin{table*}[h!]
\centering
\resizebox{\textwidth}{!}{
\begin{tabular}{llllll}
\toprule
Model & Stage & Total & Architecture & Context \\
\midrule

\href{https://huggingface.co/aisingapore/Nemotron-SEA-LION-v4.8-30B-A3B-Base}{Nemotron-SEA-LION-v4.8-30B-A3B-Base}
& CPT
& 30B
& Mamba2-Transformer Hybrid MoE
& 128K \\

\href{https://huggingface.co/aisingapore/Nemotron-SEA-LION-v4.8-30B-A3B}{Nemotron-SEA-LION-v4.8-30B-A3B}
& Post-training
& 30B
& Mamba2-Transformer Hybrid MoE
& 128K \\

\href{https://huggingface.co/aisingapore/Nemotron-SEA-LION-v4.8-120B-A12B-Base}{Nemotron-SEA-LION-v4.8-120B-A12B-Base}
& CPT
& 120B
& Mamba2-Attention Hybrid LatentMoE
& 128K \\

\href{https://huggingface.co/aisingapore/Nemotron-SEA-LION-v4.8-120B-A12B}{Nemotron-SEA-LION-v4.8-120B-A12B}
& Post-training
& 120B
& Mamba2-Attention Hybrid LatentMoE
& 128K \\

\bottomrule
\end{tabular}
}
\vspace{-2mm}
\caption{Overview of the SEA-LION-v4.8 model family. We release continued-pretrained base models and corresponding post-trained models at two computational scales.}
\vspace{-3mm}
\label{tab:sealion_v48_models}
\end{table*}

\section{Model Architecture}

SEA-LION-v4.8 consists of two complementary development tracks: a SEA-focused continued pre-training track and a post-training-focused track. Rather than assuming that the difference between continued-pretraining and post-training of the models is simply a matter of scale, we separate the two tracks to reflect distinct strategies for incorporating regional capability into large language models.
In particular, the continued pre-training track investigates adaptation at the foundation-model level, whereas the post-training track investigates how far strong existing foundation models can be specialized for SEA through intensive post-training.

\subsection{Model Variants}

As shown in Table~\ref{tab:sealion_v48_models}, \textsc{SEA-LION-v4.8} consists of four released checkpoints at two model scales: 30B parameters with 3B active parameters and 120B parameters with 12B active parameters. 
For each size, we release both a continued-pretrained base model and a corresponding post-trained model. This design allows the continued-pretraining and post-training stages to be performed independently while also providing complete end-to-end SEA-LION-v4.8 models.

\paragraph{\href{https://huggingface.co/aisingapore/Nemotron-SEA-LION-v4.8-30B-A3B-Base}{Nemotron-SEA-LION-v4.8-30B-A3B-Base}.}
Our smaller base model is initialized from
\texttt{nvidia/NVIDIA-Nemotron-3-Nano-30B-A3B-Base-BF16}.
The model contains approximately 30B total parameters with 3B active parameters and uses a Mamba2-Transformer hybrid mixture-of-experts (MoE) architecture.
To adapt the foundation Nemotron-SEA-LION-v4.8-30B-A3B-Base model toward Southeast Asian languages and contexts, we perform continued pre-training on 150B high-quality tokens spanning question answering, reasoning, code, and multilingual translation data. 
Training is performed using \href{https://github.com/NVIDIA-NeMo/Megatron-Bridge}{Megatron Bridge}.
We retain the original tokenizer without modification and preserve the model's maximum context length of 1M tokens~\footnote{However, during the post-training step, we set the context length only to 128k.}.
This checkpoint represents the output of the continued-pretraining stage before SEA-LION-v4.8 post-training.

\paragraph{\href{https://huggingface.co/aisingapore/Nemotron-SEA-LION-v4.8-30B-A3B}{Nemotron-SEA-LION-v4.8-30B-A3B}.}
The post-trained 30B-A3B model is initialized from
\textsc{Nemotron-SEA-LION-v4.8-30B-A3B-Base}.
It therefore inherits the same 30B-total/3B-active Mamba2-Transformer hybrid MoE architecture, tokenizer, and 1M-token context window of the continued-pretrained model.
We further adapt the model using the SEA-LION-v4.8 post-training framework, with online on-policy distillation (OPD) as the primary online learning objective. 
%
%
This model represents the complete 30B-A3B SEA-LION-v4.8 pipeline, combining Southeast-Asia-focused continued pre-training with large-scale online post-training.

\paragraph{\href{https://huggingface.co/aisingapore/Nemotron-SEA-LION-v4.8-120B-A12B-Base}{Nemotron-SEA-LION-v4.8-120B-A12B-Base}.}
Our larger base model is initialized from
\texttt{nvidia/NVIDIA-Nemotron-3-Super-120B-A12B-BF16}.
The model contains approximately 120B total parameters with 12B active parameters and uses NVIDIA Nemotron 3 Super's hybrid LatentMoE architecture, which interleaves Mamba-2, mixture-of-experts, and attention layers and additionally incorporates multi-token prediction.
We perform continued pre-training on 33.5B high-quality tokens covering multiple tasks and languages. Training is also performed using \href{https://github.com/NVIDIA-NeMo/Automodel}{NVIDIA NeMo AutoModel}, where we retain the original Nemotron tokenizer without vocabulary modification. 
The resulting checkpoint serves as the higher-capacity continued-pretrained model in the SEA-LION-v4.8 family.

\paragraph{\href{https://huggingface.co/aisingapore/Nemotron-SEA-LION-v4.8-120B-A12B}{Nemotron-SEA-LION-v4.8-120B-A12B}.}
The final 120B-A12B model is initialized from
\textsc{Nemotron-SEA-LION-v4.8-120B-A12B-Base} and subsequently undergoes SEA-LION-v4.8 post-training.
The model retains approximately 120B total parameters with 12B active parameters, together with the tokenizer, architecture, and 1M-token context window inherited from the continued-pretrained checkpoint.
We apply the same overall online post-training framework used for the 30B-A3B model, centered around online on-policy distillation. 
%
%
This checkpoint represents the high-capability member of the SEA-LION-v4.8 family and combines large-scale continued pre-training with online post-training at substantially greater model capacity.

\section{Continual Pre-training} \label{sec:pretraining}

For \textsc{SEA-LION-v4.8}, we perform continued pre-training (CPT) on two NVIDIA Nemotron 3 foundation models at different scales: \href{https://huggingface.co/nvidia/NVIDIA-Nemotron-3-Nano-30B-A3B-Base-BF16}{nvidia/NVIDIA-Nemotron-3-Nano-30B-A3B-Base-BF16} and \href{https://huggingface.co/nvidia/NVIDIA-Nemotron-3-Super-120B-A12B-BF16}{nvidia/NVIDIA-Nemotron-3-Super-120B-A12B-BF16}.
The resulting checkpoints are released as
\textsc{Nemotron-SEA-LION-v4.8-30B-A3B-Base} and
\textsc{Nemotron-SEA-LION-v4.8-120B-A12B-Base}, respectively.

Our CPT stage is designed to strengthen the representation of Southeast Asian languages while preserving the reasoning, coding, and general capabilities inherited from the original Nemotron 3 models.
Rather than relying primarily on generic web text, we emphasize structured and capability-oriented data, including question answering, chain-of-thought reasoning, mathematical and scientific reasoning, code, and bilingual parallel data.
%
%
The 30B-A3B model is trained using Megatron Bridge, whereas the 120B-A12B model is trained using NVIDIA NeMo AutoModel.
The training data and CPT strategy are described below.

\subsection{Pre-training Data}

\begin{figure}[h!]
    \centering
    \includegraphics[width=\linewidth]{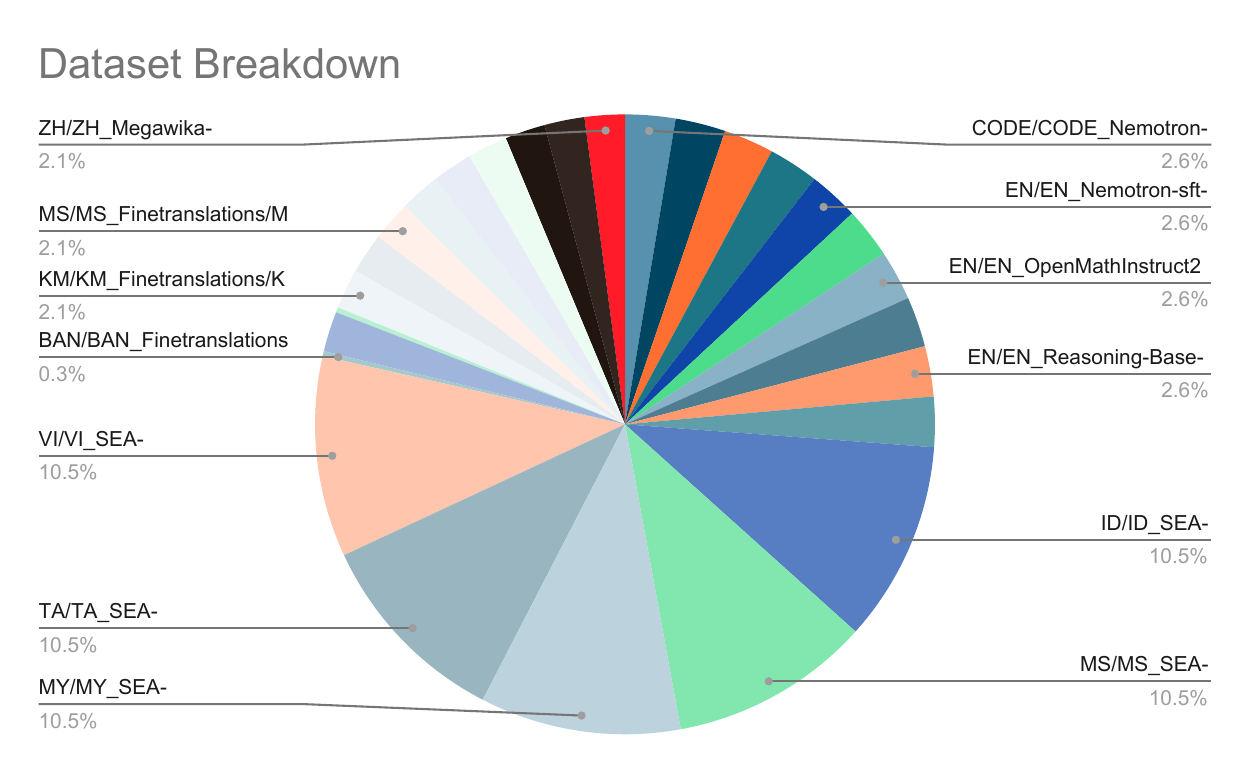}
    \vspace{-3mm}
    \caption{
    Distribution of the 150B-token continued pre-training corpus.
    The mixture combines SEA-focused instruction data with reasoning, code, and bilingual parallel data.
    }
    \vspace{-3mm}
    \label{fig:cpt_distribution}
\end{figure}

We construct a 150B-token CPT corpus consisting of Southeast Asian instruction data, reasoning-oriented data, code, and bilingual parallel data.
The corpus is designed to increase exposure to Southeast Asian languages while retaining the general reasoning and knowledge capabilities of the original Nemotron 3 foundation models.
The training data can be broadly divided into three groups.
In addition, Table~\ref{tab:cpt_data} summarizes the main components of the CPT mixture, while Figure~\ref{fig:cpt_distribution} shows the overall data distribution.

\noindent
\textbf{Southeast Asian data.}
A substantial portion of the training mixture consists of Southeast Asian data covering Balinese, Burmese, Indonesian, Javanese, Khmer, Lao, Malay, Sundanese, Tamil, Thai, Vietnamese, and other regional languages.
High-weight \href{https://huggingface.co/datasets/aisingapore/SEA-Instruct-2602}{SEA-Instruct} data are included for Burmese, Indonesian, Malay, Tamil, and Vietnamese, while additional language coverage is provided through bilingual parallel corpora.

\noindent
\textbf{Reasoning and code data.}
To preserve and strengthen the reasoning capabilities of the underlying foundation models during CPT, we include English and multilingual datasets covering code, mathematics, scientific reasoning, general reasoning, question answering, and chain-of-thought supervision.
The mixture additionally includes Thai medical reasoning data.

\noindent
\textbf{Parallel multilingual data.}
We include bilingual parallel data in which a Southeast Asian language is paired with English.
For each pair, both translation directions are represented by alternating the source and target languages across examples.
This exposes the model to both SEA-to-English and English-to-SEA mappings and provides an explicit cross-lingual training signal.

\begin{table*}[h!]
\centering
\begin{tabular}{lll}
\toprule
Data Group & Languages / Domains & Sampling Weight \\
\midrule
SEA-Instruct & Indonesian & 10 \\
& Malay & 10 \\
& Burmese & 10 \\
& Tamil & 10 \\
& Vietnamese & 10 \\
\midrule
Reasoning / Code & Code & 2.5 per dataset \\
& Mathematics & 2.5 per dataset \\
& Scientific reasoning & 2.5 per dataset \\
& General reasoning / CoT & 2.5 per dataset \\
& Thai medical reasoning & 2.5 \\
\midrule
Parallel Data & Indonesian & 2 \\
& Khmer & 2 \\
& Lao & 2 \\
& Malay & 2 \\
& Burmese & 2 \\
& Tamil & 2 \\
& Thai & 2 \\
& Vietnamese & 2 \\
& Chinese & 2 \\
& Bengali & 0.25 \\
& Javanese & 0.25 \\
& Sundanese & 0.25 \\
\bottomrule
\end{tabular}
\vspace{-2mm}
\caption{
Overview of the continued pre-training data mixture used for
\textsc{SEA-LION-v4.8}. Sampling weights determine the relative contribution of each component to the 150B-token CPT corpus. 
}
\vspace{-3mm}
\label{tab:cpt_data}
\end{table*}

\subsection{Tokenizer}

We retain the original tokenizer of each Nemotron 3 foundation model without modification.
Retaining the original tokenizer preserves compatibility with the pretrained embedding and output layers and avoids introducing additional architectural changes during CPT.
However, this decision also exposes an important limitation of continued pre-training: additional language data cannot fully compensate for inefficient tokenization.

\begin{figure}[h!]
    \centering
    \includegraphics[width=\linewidth]{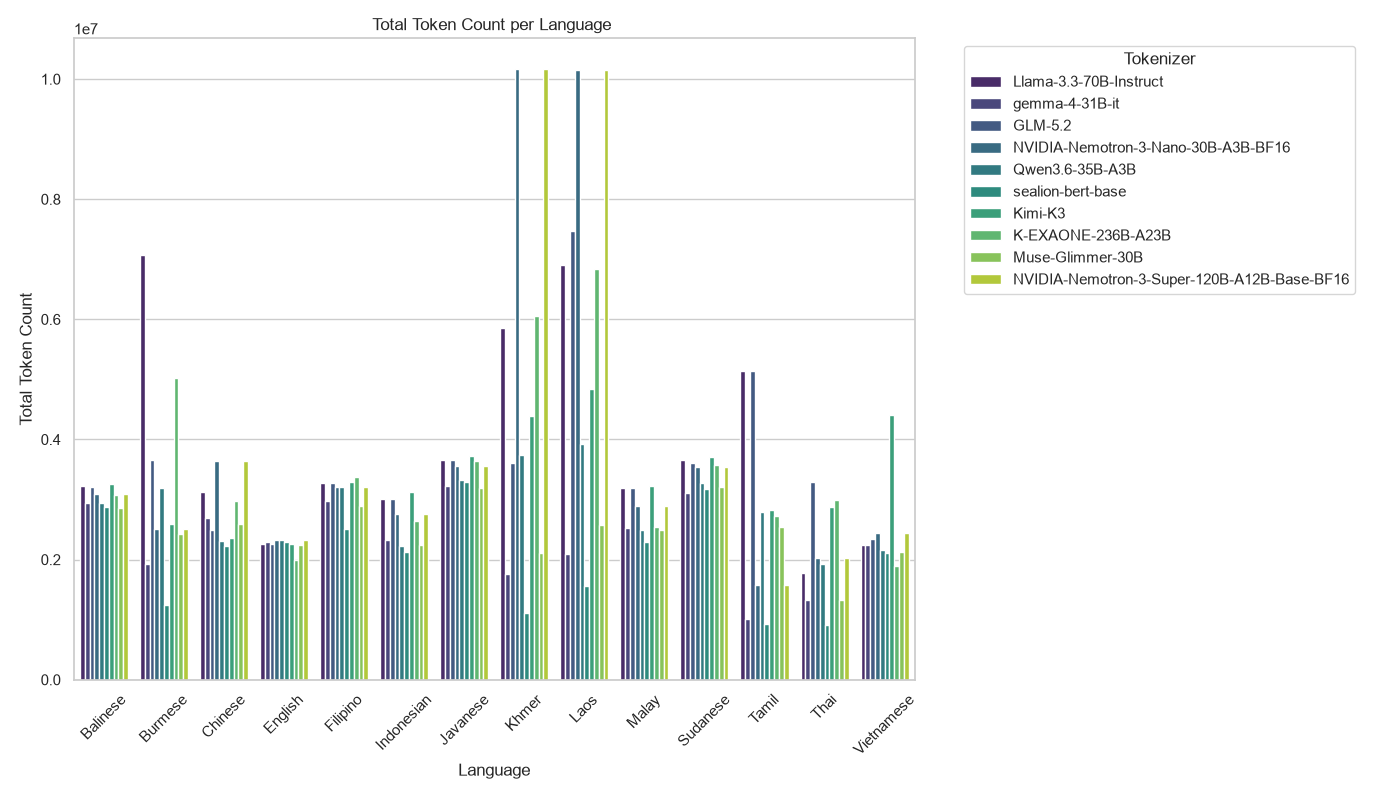}
    \vspace{-5mm}
    \caption{
    Comparison of total token counts produced by different tokenizers across languages.
    Lower token counts indicate more compact text representations.
    }
    \vspace{-5mm}
    \label{fig:token_distribution}
\end{figure}

\begin{figure}[h!]
    \centering
    \includegraphics[width=\linewidth]{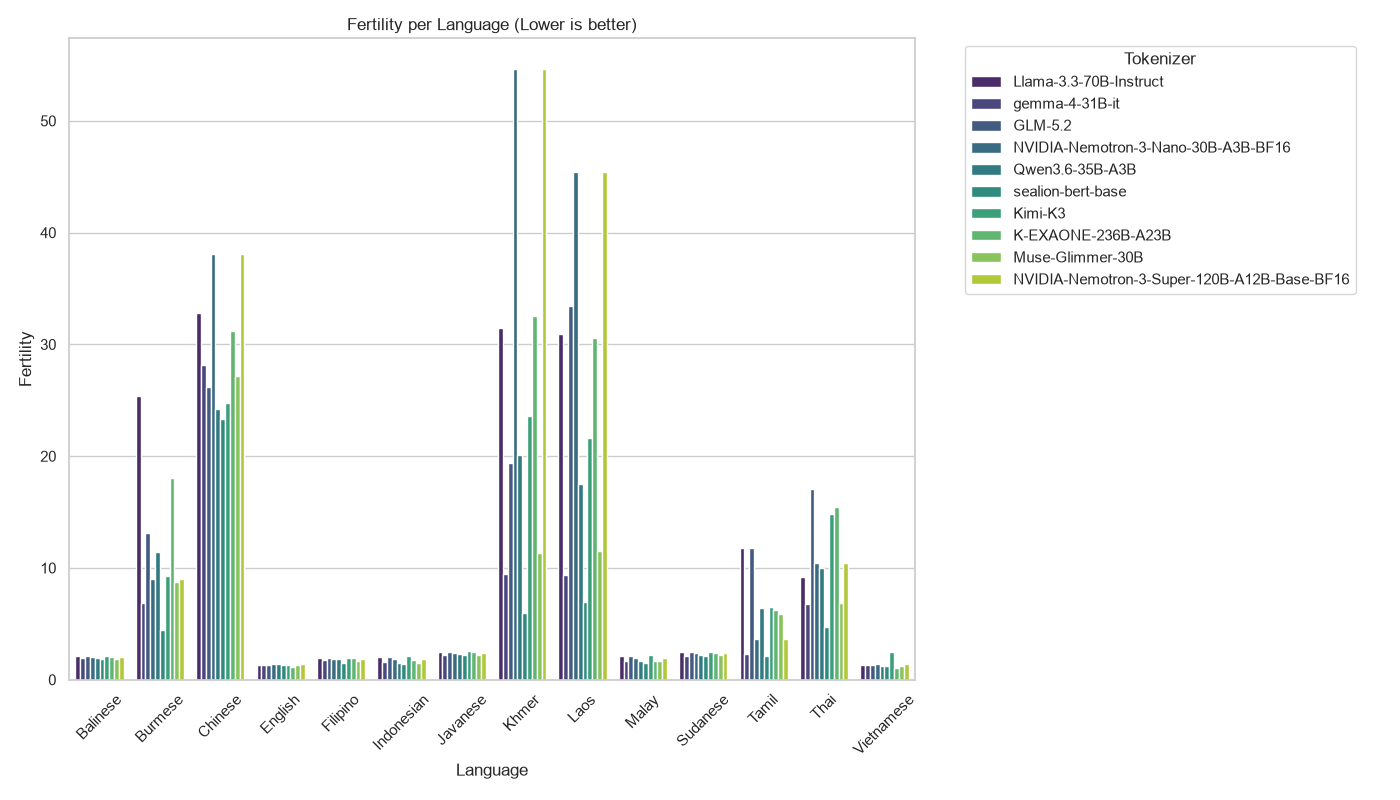}
    \vspace{-5mm}
    \caption{
    Tokenizer fertility across languages.
    Lower fertility indicates that fewer tokens are required to represent the same text.
    }
    \vspace{-5mm}
    \label{fig:token_score}
\end{figure}

As shown in Figures~\ref{fig:token_distribution} and~\ref{fig:token_score}, our tokenizer analysis for the 30B-A3B model reveals substantial differences in tokenization efficiency across Southeast Asian languages.
The Nemotron 3 tokenizer produces substantially more tokens for several regional languages than tokenizers designed with stronger SEA coverage.
In particular, for Khmer, Lao, and Tamil, the tokenizer produces approximately five times as many tokens as the previous SEA-LION tokenizer.
This fragmentation reduces the amount of linguistic information represented within a fixed context and increases the effective sequence length of the corresponding language.
Consequently, a language may receive substantial CPT exposure while still benefiting less from training because its text is represented inefficiently.

%

\subsection{Training Recipe}

Our CPT recipe follows four main design choices.
We use a maximum learning rate of $3\times10^{-5}$, following prior work showing that relatively high learning rates can be effective during continued pre-training~\citep{DBLP:journals/corr/abs-2308-04014}.
Rather than relying primarily on generic unstructured text, our CPT mixture places substantial weight on question answering, chain-of-thought, mathematical, scientific, and general reasoning data.
The objective is to improve regional language exposure while maintaining a strong capability foundation for the subsequent post-training stage~\citep{DBLP:journals/corr/abs-2510-03264,DBLP:conf/acl/JiangSSRZNLY024}.
Moreover, we use bilingual parallel data to provide an explicit cross-lingual learning signal.
By presenting both SEA-to-English and English-to-SEA examples, the model is encouraged to align representations across languages rather than learning each regional language independently~\citep{DBLP:journals/corr/abs-2602-02266}.
Lastly, we adopt a WSD learning-rate schedule but omit the final decay stage.
Our motivation is to avoid aggressively converging the CPT checkpoint before post-training and to preserve sufficient model plasticity for subsequent adaptation~\citep{DBLP:journals/corr/abs-2603-16127,DBLP:journals/corr/abs-2511-18903}.
Lastly, we set the hyper-parameter and configuration as shown in Table~\ref{tab:cpt_setup}.

\begin{table*}[h!]
\centering
\resizebox{\textwidth}{!}{
\begin{tabular}{lll}
\toprule
Configuration & 30B-A3B & 120B-A12B \\
\midrule
Initialization
& Nemotron 3 Nano 30B-A3B Base
& Nemotron 3 Super 120B-A12B Base \\

Training tokens
& 150B
& 33.5B \\

Training framework
& \href{https://github.com/NVIDIA-NeMo/Megatron-Bridge}{Megatron Bridge}
& \href{https://github.com/NVIDIA-NeMo/Automodel}{NeMo AutoModel} \\

Tokenizer
& Original Nemotron tokenizer
& Original Nemotron tokenizer \\

Maximum learning rate
& $3\times10^{-5}$
& $1\times10^{-5}$ \\

LR scheduler 
& WSD without final decay
& WSD without final decay \\

Maximum sequence length
& 8192
& 8192 \\

Optimizer
& AdamW
& AdamW \\

Global batch size
& 1024
& 1024 \\

Warmup
& 5\%
& 5\% \\

Precision
& BF16
& BF16 \\

Training hardware
& H200
& H200 \\

Number of GPUs
& 32
& 32 \\

Training time
& 114.23 hrs
& 186.65 hrs \\
\bottomrule
\end{tabular}}
\vspace{-2mm}
\caption{
Continued pre-training configurations for the 30B-A3B and 120B-A12B base models.
}
\vspace{-3mm}
\label{tab:cpt_setup}
\end{table*}

\subsection{Limitations of Continued Pre-training}

Our CPT experiments expose two important limitations of adapting existing foundation models to Southeast Asian languages.

First, \textbf{tokenization can impose a fundamental bottleneck on language adaptation}.
Although we substantially increase the amount of Southeast Asian data seen during CPT, the underlying Nemotron tokenizer is retained.
For several regional scripts, the tokenizer represents text much less efficiently than tokenizers explicitly optimized for Southeast Asian languages.
%
This result indicates that regional adaptation cannot be treated solely as a data-scaling problem.
If a target language is poorly represented in the tokenizer vocabulary, increasing the amount of training data may provide diminishing returns.
Addressing this limitation may require vocabulary expansion, tokenizer adaptation, or language-aware initialization of newly introduced embeddings.
We leave these directions for future work because modifying the tokenizer would also require changes to pretrained embedding and output parameters.

Second, \textbf{Filipino is not explicitly represented in the current CPT corpus}.
Although Filipino is relevant to our downstream Southeast Asian evaluation and deployment setting, no dedicated Filipino training component is included in the present CPT mixture.
Any improvement on Filipino therefore arises from cross-lingual transfer from other languages and from improvements in the model's broader multilingual capabilities rather than direct Filipino-specific continued pre-training.
This distinction is important because the failure modes are different.
For Tamil and Burmese, training data are available, but the model is constrained by representation and tokenization quality.
For Filipino, the primary limitation is direct data coverage.
Future iterations should incorporate high-quality Filipino corpora and investigate whether explicit Filipino CPT produces gains beyond those obtained through cross-lingual transfer.

Together, these cases show that successful regional continued pre-training depends not only on the total amount of Southeast Asian data, but also on whether individual target languages are adequately represented in both the \emph{training distribution} and the model's \emph{tokenization space}.

\section{Post-training}
\label{sec:posttraining}

Starting from the continued-pretrained \textsc{Nemotron-SEA-LION-v4.8-30B-A3B-Base} and \textsc{Nemotron-SEA-LION-v4.8-120B-A12B-Base} checkpoints, we further adapt both model sizes using a unified online post-training framework.
%
%
In particular, we iteratively enhance model's capabilities through self-generated exploratory experiences using a strong signal to guide the model towards the right direction. 
%
%
To generate a strong signal, we produce it from a stronger teacher model that scoring the original response of the student model, so that the student learns at each token position what the teacher would have chosen instead as a signal, so it can make the same generations as the stronger model.

%
%

Moreover, we also design a separate environment interaction, token-exact trajectory capture, and model optimization, so that each service is decoupled and scaled individually and easily extendable. 
Agent trajectories are inherently unpredictable, long-tailed, and difficult to attribute rewards to, especially in online training, where the agent needs its answer at serving latency, but training evidence could trail behind; the scaling has to be decoupled to be able to optimize. 
Since every subsequent turn re-submits the whole history, it is especially important to identify the correct session and stitch it onto an existing conversation or forked incase of compaction or change in history.
This allows environments to scale and continuously produce new interactions asynchronously while the trainer consumes previously completed trajectories, creating a continuously evolving training distribution.

\subsection{Post-training Data}

Unlike continued pre-training, the main contribution of the SEA-LION-v4.8 post-training stage is not a new static data distribution, but rather focuses on the unified training pipeline that is modular and decoupled.
Instead, the training stream combines high-quality offline supervised examples with agentic trajectories generated dynamically by the model during training. The data is being consumed using 1 environment that acts like a data feeder to either directly consume the offline supervised data as training samples or to pass it to agent harnesses to create agentic trajectories. The agents are scaled dynamically based on the load on the inference server.
Therefore, we divide the post-training data into two broad sources: \emph{offline supervised data} and \emph{online interaction data}.

\noindent
\textbf{Offline supervised data.}
The unified training pipeline can directly consume conventional supervised fine-tuning examples.
These examples are a subset from our aisingapore/SEA-Instruct-2602 dataset and cover a wide range of South East Asian Langugages namely Indonesian, Vietnamese, Thai, Filipino, Tamil, Tagalog, Malay and Burmese and English, all in equal proportions. This include instruction-following supervision datasets and are injected into the same training pipeline as online-generated trajectories, but scored with a different SFT loss instead.

\noindent
\textbf{Online interaction data.}
The input prompts are the same as the supervised fine tuned examples. 
The second source of post-training data consists of trajectories generated dynamically by the model through a large collection of environments and agent harnesses.
Each environment defines an interaction protocol, a task that interacts with an agent harness, relying on the model currently being trained to produce the language-model responses.
The environments cover heterogeneous interaction patterns, including direct instruction-following tasks, multi-turn interactions, agentic tasks, and other structured scenarios.
Some environments also include persona-based debate or multi-agent-style interactions.
Unlike conventional offline distillation, these trajectories are not generated once and stored as a fixed dataset.
Instead, the current student model generates new trajectories throughout training.
As the model parameters change, the distribution of generated responses changes accordingly, allowing the training data to evolve together with the student policy.
To meet the scale of the training system, the agent environment orchestrator supports approximately 3,000 simultaneously active agent instances and processes more than 100,000 interaction sessions over the course of training.
A key property of this design is that the agent instances themselves contain no independent language models and use the model in training as the inference engine.
A single model checkpoint powers all active agents, while the individual environments determine the task, environment state, interaction structure, and other task-specific behavior for each agent harness.

\subsection{Training Recipe}

\begin{figure}[h!]
    \centering
    \includegraphics[width=\linewidth]{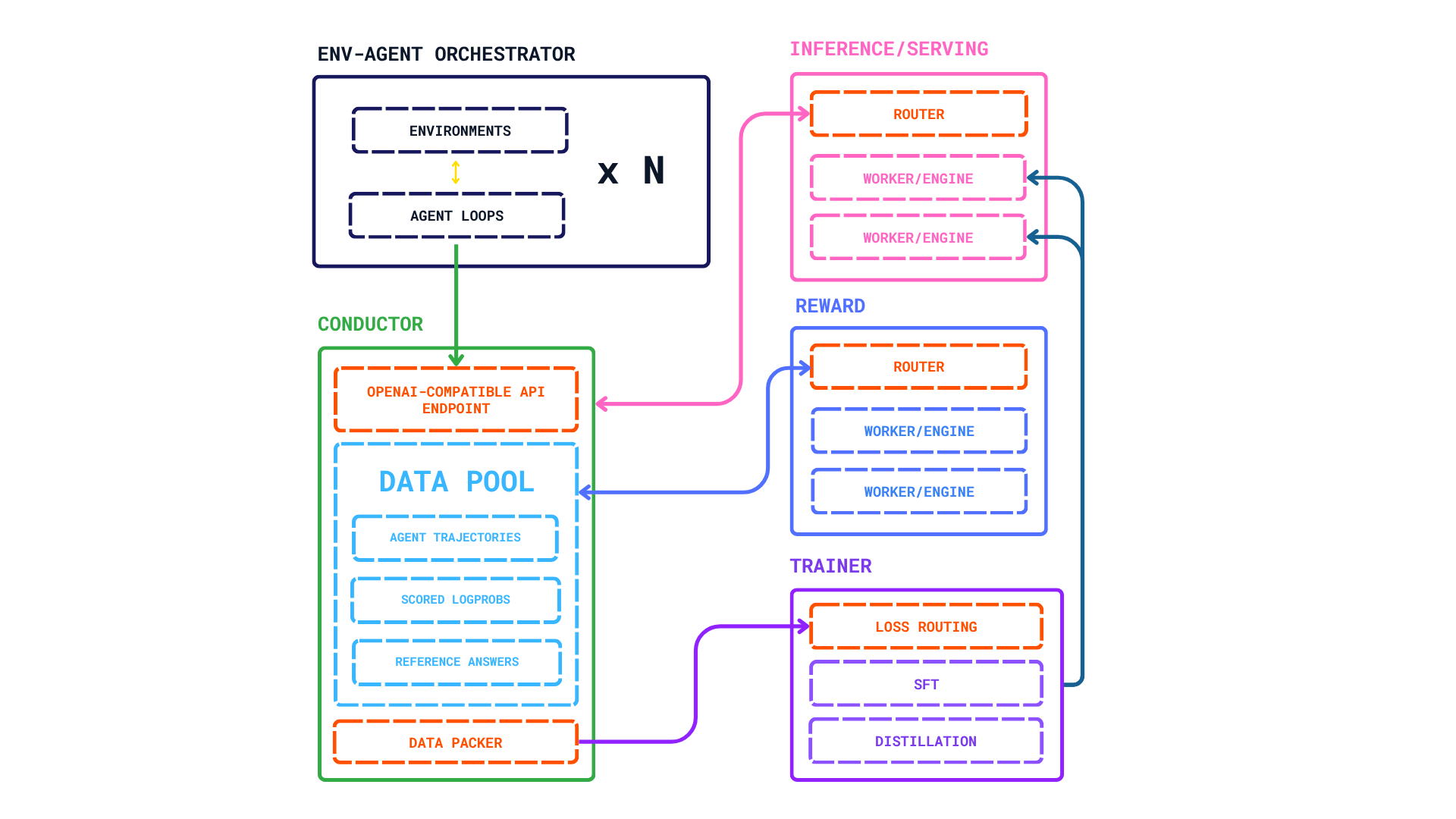}
    \vspace{-5mm}
    \caption{
    An overview of the post-training framework.
    }
    \vspace{-3mm}
    \label{fig:post_training_overview}
\end{figure}

As shown in Figure~\ref{fig:post_training_overview}, the SEA-LION-v4.8 post-training pipeline consists of five principal components: \emph{environment-agent orchestrator}, \emph{Conductor and training database}, \emph{inference service}, \emph{reward service}, and \emph{trainer}.
Together, these components form an asynchronous, modular, fully decoupled agent-oriented design training pipeline that separates rollout, reward engines from training engines for generating interactions, recording model behavior, and continuously updating model parameters.
%

%
%
%

\noindent
\textbf{Environment-agent orchestrator.} The environment-agent orchestrator dynamically defines and scales the containers needed, based on watching the conductor endpoint. 
It manages a collection of different environments and agent harnesses that expose the model to different tasks, environments, and interaction structures.
Each environment container defines the objective and enforces how the tasks should be carried out. The agent harness does not contain its own language model, 
Whenever a model response is required, the harness sends the corresponding request through the training system to the model currently being optimized.
This separation between the model and the task harnesses allows a single evolving model to participate simultaneously in thousands of heterogeneous interactions.
The orchestrator is able to scale dynamically and can sustain 3,000 agent instances active concurrently, depending on what the training requires.
The orchestrator can additionally inject conventional supervised examples into the training stream, allowing offline SFT data and online interaction data to be handled within the same training infrastructure.

\noindent
\textbf{Conductor and token-exact data capture.}
All training interactions pass through \emph{Conductor}, which acts as the interface between the task environments and the training system.
A central design principle is to train on the exact model behavior that occurred during interaction.
Therefore, the Conductor records model responses and the associated training information before the responses are returned to the corresponding environments.
The resulting records are stored in a central training database that serves as the system of record for the post-training process, where the database preserves the exact token sequences and associated metadata required for subsequent training.
This token-exact capture avoids reconstructing model trajectories after generation and ensures that the trainer optimizes the model using the behavior that was actually produced.
The separation between interaction generation and optimization also enables asynchronous execution: environments can continue producing new trajectories while the trainer independently consumes previously completed records.

\noindent
\textbf{Serving and reward fleet.}
We used a modified version of \href{https://github.com/sgl-project/sglang}{SGLang} for both serving and reward.
Incoming requests are preferentially routed to target workers using session ID affinity. If an existing session is not found, routing defaults to the SGLang worker holding the matching key-value (KV) cache. 
However, to prevent bottleneck, this prefix-cache affinity is overridden in favor of load balancing if the target worker exceeds safe capacity, re-routing the request to the lowest-workload instance. 
Within the reward fleet, top-$k$ teacher outputs are compressed and binarized before sending over. 
We only score and materialize the unmasked token positions that we want to train on for each sequence. 
The serving and reward fleets share a unified compute pool and dynamically scale worker instances up or down based on backlogs in each respective queue. 
Model weights are synchronized continuously via step-wise updates from the trainer while preserving stale KV cache states, and all generated tokens are versioned individually at the token level to guarantee tracking accuracy across model iterations.

\noindent
\textbf{Asynchronous trainer.}
We used a modified version of \href{https://github.com/NVIDIA-NeMo/Automodel}{AutoModel}. The trainer continuously consumes training records from the central database. We train tokens based on staleness up to 32. Tokens outside of this window are masked. When we consume training records, they are also packed to the maximum sequence length. We ensure to keep rollouts and the off policy reference answer from the same prompt together into the same batch.
Instead of first generating an entire dataset and subsequently launching a separate optimization stage, trajectory generation and model training proceed concurrently.
Once an interaction becomes available, it can be selected for optimization without waiting for the overall data-collection process to finish. 
The resulting training loop is:

\begin{quote}
environment interaction
$\rightarrow$
student response
$\rightarrow$
trajectory capture
$\rightarrow$
training
$\rightarrow$
updated student
$\rightarrow$
new interactions.
\end{quote}

The trainer can consume both offline supervised examples and online-generated trajectories through a unified interface.
This enables conventional supervised fine-tuning and online distillation to be interleaved during the same post-training run.

\noindent \textbf{Online on-policy distillation.}
The primary online learning method used in SEA-LION-v4.8 is
\emph{online on-policy distillation over the teacher's top k} (OPD).
Given an input or an environment state, the current student model first generates a response using its current policy.
A stronger teacher model then provides supervision over the student's generated trajectory.
The student is subsequently optimized using a KL-based distillation objective.
The important distinction from conventional offline knowledge distillation is that the training trajectories originate from the \emph{current student policy}.
They are not generated in advance by a fixed teacher and then reused throughout training.
Let the student distribution at training step $t$ be
$p_{\theta_t}$.
The online interaction process produces trajectories

\[
\tau_t \sim p_{\theta_t}.
\]

Teacher supervision is then evaluated on $\tau_t$, and the resulting signal is used to update the student,

\[
\theta_t
\rightarrow
\theta_{t+1}.
\]

Consequently, the training distribution evolves together with the student:

\[
\tau_{t+1} \sim p_{\theta_{t+1}}.
\]

This allows the teacher to provide supervision specifically on behaviors that the current student is likely to produce, rather than only on trajectories sampled from a fixed offline distribution.
Both for the 30B-A3B and model and 120B-A12B model, the teacher model is \textsc{\href{https://huggingface.co/RedHatAI/NVIDIA-Nemotron-3-Ultra-550B-A55B-FP8-dynamic}{RedHatAI/NVIDIA-Nemotron-3-Ultra-550B-A55B-FP8-dynamic}}. 

%
%

\noindent \textbf{Joint SFT and OPD training.}
We do not treat supervised fine-tuning and online distillation as completely independent sequential stages.
The same trainer can consume offline supervised examples and online-generated OPD trajectories.
For an SFT example, the model is optimized using the corresponding supervised objective.
For an online trajectory, the model instead receives teacher supervision through the OPD objective.
The effective training stream can therefore be written as

\[
\mathcal{D}_{\mathrm{train}}
=
\mathcal{D}_{\mathrm{SFT}}
\cup
\mathcal{D}_{\mathrm{OPD}},
\]

%
This design allows curated supervised data to provide a stable instruction-following foundation while OPD continually exposes the model to trajectories generated from its own evolving policy.

%

\noindent \textbf{Training configuration.}
Table~\ref{tab:posttraining_setup} summarizes the post-training configurations of the two released SEA-LION-v4.8 models.
After tuning with OPD as the primary objective together with a combination of SFT loss, we perform merging with the nvidia/NVIDIA-Nemotron-3.5-Lightning-30B-A3B-BF16 and nvidia/NVIDIA-Nemotron-3-Super-120B-A12B-BF16 respectively. 
For the 30B model, we tuned it for 1200 Steps and 1600 steps for the 120B model.

\begin{table*}[h!]
\centering
\resizebox{\textwidth}{!}{
\begin{tabular}{lll}
\toprule
Configuration
& 30B-A3B
& 120B-A12B \\
\midrule

Initialization
& \textsc{Nemotron-SEA-LION-v4.8-30B-A3B-Base}
& \textsc{Nemotron-SEA-LION-v4.8-120B-A12B-Base} \\

Post-training methods
& SFT + online OPD
& SFT + online OPD \\

Teacher model
& RedHatAI/NVIDIA-Nemotron-3-Ultra-550B-A55B-FP8-dynamic
& RedHatAI/NVIDIA-Nemotron-3-Ultra-550B-A55B-FP8-dynamic \\

Distillation objective
& Reverse KL
& Reverse KL \\

Concurrent agent instances
& $\sim$3,000 at system scale
& $\sim$3,000 at system scale \\

Total interaction sessions
& $\ge$100,000 across training 
& $\ge$100,000 across training \\


Maximum sequence length
& 128,000
& 128,000 \\

Optimizer
& AdamW
& AdamW \\

Maximum learning rate
& 1e-5
& 1e-5 \\

Global batch size
& 24
& 24 \\

Training hardware
& H200
& H200 \\

Training /No. of GPUs
& 16 
& 48 \\

Reward /No. of GPUs
& 48
& 48 \\

Serving /No. of GPUs
& 8
& 16 \\

Training duration
& 12 hours
& 24 hours \\

Parallelism
& HSDP x1 (FSDP, DP2, EP8, CP8, PP1) 
& HSDP x3 (FSDP, DP1, EP8, CP8, PP2, Interleaved 1f1b)\\

\bottomrule
\end{tabular}}
\caption{
Post-training configuration for
\textsc{Nemotron-SEA-LION-v4.8-30B-A3B} and
\textsc{Nemotron-SEA-LION-v4.8-120B-A12B}.
Both models are trained using a shared framework combining supervised fine-tuning with online on-policy distillation. *Reward and Serving is only at the initialized values but changes dynamically during training.
}
\label{tab:posttraining_setup}
\end{table*}

\section{Evaluation: SEA-HELM}
The updated version of SEA-HELM~\citep{susanto-etal-2025-sea} (Southeast Asian Holistic Evaluation of Language Models) was used to evaluate the SEA-LION-v4.8 models. SEA-HELM is a holistic evaluation suite designed to assess the linguistic and cultural competencies of LLMs with respect to Southeast Asia. It was introduced to address the lack of a benchmark constructed with substantive community participation that provides a comprehensive and culturally representative coverage of Southeast Asian languages. 

Several design commitments have defined the suite since its inception and were retained in this updated version. 
First, to avoid translationese and cultural erasure associated with machine-translated benchmarks, datasets were curated to be natively written wherever possible, or otherwise carefully translated and localized in collaboration with native speakers. 
Second, to ensure the quality of datasets, native speakers of the target languages participate during each stage of dataset planning and construction, and task prompts are written in the target language itself. 
Third, results are normalized against random-baseline performance and aggregated hierarchically, from task to competency to language to an overall Southeast Asian average, and are presented through a publicly accessible leaderboard\footnote{\url{https://leaderboard.sea-lion.ai/}} that offers an overall, per-language, and per-task views.
This updated suite preserved the evaluation philosophy summarized above while expanding language coverage, broadening the treatment of culture and knowledge, and strengthening the statistical methodology underpinning reported results. 
The full details of each task are available in \href{https://leaderboard.sea-lion.ai/tasks/language}{the SEA-HELM leaderboard}.

\subsection{Language Coverage}
In this update, two additional languages, Malay and Burmese~\citep{aung2026burmesesanburmesenlpbenchmark}, have been added. This builds on the existing coverage of the Filipino, Indonesian, Tamil, Thai, and Vietnamese languages, bringing the total number of SEA languages supported to 7.

\subsection{Additional Evaluation Dimensions}
Cultural capabilities were evaluated primarily using SEA-NLI and, additionally, for Filipino, KALAHI. SEA-NLI~\citep{chomphooyod2026seanlinaturallanguageinference} assesses the model's ability to draw cultural relationships between specific entities in the native language. This evaluates the model's cultural understanding and reasoning. KALAHI~\citep{montalan2025kalahihandcraftedgrassrootscultural}, a participatory dataset that was designed with the help of native speakers of Filipino from the Philippines, assesses whether models can select culturally appropriate responses to situations that members of the target community plausibly encounter.  This release further extends this evaluation task by not providing any options for models to choose from. This forces the models to generate the culturally appropriate response independently. 
These responses were then graded using LLM-as-a-Judge with targeted criteria for each prompt.
In addition to cultural knowledge, models were also evaluated on their general world knowledge in the various Southeast Asian languages via the addition of Global MMLU-Lite \citep{singh-etal-2025-global} and Thai Exam \citep{pipatanakul2023typhoon}.

The linguistic capabilities of the models were evaluated using  LINDSEA (Linguistic Diagnostics for Southeast Asian languages) \citep{leong2023bhasaholisticsoutheastasian}, a hand-crafted dataset constructed by linguists in collaboration with native speakers to probe models' grammatical and linguistic understanding at a fine-grained level. Similar to KALAHI, this release introduced a generative variant of the linguistic diagnostic task. Models were required to generate sentences that exhibited the linguistic phenomena, and these tests whether models know and, more importantly, are able to apply these linguistic rules.

The ability of models to be able to distinguish between safe and unsafe prompts is vital to ensure that models are ready for interacting with a general audience. To that end, SEA-HELM now includes SEA-SafeguardBench~\citep{tasawong-etal-2026-sea-safeguardbench}, a task that diagnoses whether models are able to identify culturally inappropriate prompts and responses.

\subsection{Evaluation Methodology}

To account for the statistical nature of the models and the reality that deployed models are often run with temperatures higher than 0, we chose to forgo the single-run methodology of the original release. Instead, we employed eight independent runs per model using the default generation configurations defined in each model's configuration. For each prompt, we calculate the average score across the eight independent runs to account for the run-to-run variance of the generated answer.
Additionally, bootstrapping with resampling across the prompts for each task was performed. A total of 2000 bootstraps were performed, and the 95\% confidence interval was calculated from the bootstraps.

\section{Main Results}

\subsection{Language Performance}

\noindent
\textbf{Goal.}
The goal of this experiment is to evaluate whether \textsc{SEA-LION-v4.8} improves language-specific performance across Southeast Asian languages, rather than relying only on an aggregate regional score.
%
%
We report SEA-HELM results separately for Burmese (MY), Filipino (TL), Indonesian (ID), Malay (MS), Tamil (TA), Thai (TH), and Vietnamese (VI), and compare the 30B-A3B and 120B-A12B \textsc{SEA-LION-v4.8} models against their corresponding Nemotron 3 baselines.
The live SEA-HELM leaderboard is available at \url{https://leaderboard.sea-lion.ai/}.

\noindent
\textbf{Results.}
As shown in Table~\ref{tab:sealion-compact}, \textsc{SEA-LION-v4.8} consistently improves over the corresponding Nemotron 3 models at both model scales.
The 30B-A3B model improves the overall SEA score from 46.89 to 51.57, while the 120B-A12B model shows a substantially larger improvement from 49.30 to 63.44.
This indicates that the SEA-oriented continued pre-training and post-training pipeline provides consistent gains beyond the original Nemotron 3 models, with the larger 120B-A12B model benefiting more strongly overall.

At the language level, improvements are observed across all seven evaluated SEA languages.
For the 30B-A3B model, the largest gain is observed in Tamil, improving from 22.50 to 33.14, followed by Burmese from 3.23 to 10.61.
The model also improves Filipino, Indonesian, Malay, Thai, and Vietnamese by smaller but consistent margins.
The gains become substantially more pronounced at the 120B-A12B scale: Burmese improves from 4.98 to 31.35, Tamil from 21.83 to 56.35, Thai from 56.61 to 68.66, and Vietnamese from 55.31 to 70.36.
These results suggest that increasing model capacity together with SEA-oriented adaptation is particularly beneficial for languages that are comparatively weak in the original Nemotron 3 models.

Despite these improvements, performance remains uneven across languages.
In particular, Burmese and Tamil remain more challenging than Indonesian, Malay, Filipino, Thai, and Vietnamese in absolute performance, especially for the 30B-A3B model.
This is consistent with our tokenizer analysis in Section~\ref{sec:pretraining}, which shows that the original Nemotron tokenizer provides less efficient representations for several Southeast Asian scripts.
For such languages, additional CPT data can substantially improve performance, but tokenizer inefficiency may still constrain how effectively the model learns from the available data.
This suggests that further improvements may require not only additional language-specific training data, but also better tokenizer support for underrepresented Southeast Asian scripts.

\begin{table}[h!]
\centering
\resizebox{\textwidth}{!}{
\begin{tabular}{l l r *{8}{r}}
\toprule
\textbf{Model} & \textbf{Org.} & \textbf{Size (B)} &
\textbf{SEA} & \textbf{MY} & \textbf{TL} & \textbf{ID} &
\textbf{MS} & \textbf{TA} & \textbf{TH} & \textbf{VI} \\
\midrule 

\multicolumn{11}{l}{\textit{Models $\leq$ 35B}} \\

Nemotron 3 Nano & NVIDIA & 30 & 46.89 & 3.23 & 55.53 & 65.36 & 58.00 & 22.50 & 62.90 & 60.72 \\
Nemotron 3.5 Lightning & NVIDIA & 30 & 46.06 & 3.79 & 58.69 & 64.34 & 58.76 & 17.23 & 58.66 & 60.97 \\
SEA-LION v4.8 (Nemotron)   & AISG+NVidia   & 30  & 51.57 & 10.61 & 61.82 & 65.87 & 62.09 & 33.14 & 62.60 & 64.86 \\

\midrule
\multicolumn{11}{l}{\textit{Models $>$ 35B}} \\
Nemotron 3 Super & NVIDIA & 120 & 49.30 & 4.98 & 66.92 & 70.78 & 68.66 & 21.83 & 56.61 & 55.31 \\
SEA-LION v4.8 (Nemotron)   & AISG+NVidia   & 120 & 63.44 & 31.35 & 71.01 & 73.10 & {73.25} & 56.35 & {68.66} & {70.36} \\

\bottomrule
\end{tabular}}
\caption{SEA-HELM scores by language (point estimates). Note that the scores were gathered on September 15, 2026. The score in the live leaderboard might be changed.}
\label{tab:sealion-compact}
\end{table}

\subsection{Task Performance}

\noindent
\textbf{Goal.}
While the language-level analysis measures how well \textsc{SEA-LION-v4.8} performs across individual Southeast Asian languages, it does not reveal \emph{which capabilities} are improved by our training pipeline.
Thus, we evaluate the models across eight SEA-HELM capability categories: Cultural, Instruction Following, Knowledge, Multi-turn, Natural Language Generation (NLG), Natural Language Reasoning (NLR), Natural Language Understanding (NLU), and Safety.
This capability-level analysis allows us to identify which skills benefit most from continued pre-training and post-training, and where the adapted models remain comparable to or weaker than their corresponding Nemotron baselines.
We report results separately for the 30B-A3B and 120B-A12B models across seven Southeast Asian languages.

\begin{table*}[h!]
\centering
\small
\setlength{\tabcolsep}{5pt}
\scalebox{0.8}{
\begin{tabular}{lrrrrrrr}
\toprule
\textbf{Model} &
\textbf{MY} & \textbf{TL} & \textbf{ID} & \textbf{MS} &
\textbf{TA} & \textbf{TH} & \textbf{VI} \\
\midrule

\multicolumn{8}{l}{\textbf{Cultural}} \\
\multicolumn{8}{l}{\textit{Models $\leq$ 35B}} \\
Nemotron 3 Nano
    & 0.00 & 52.23 & 56.10 & 31.86 & 10.16 & 50.71 & 39.48 \\
Nemotron 3.5 Lightning
    & 0.00 & 64.29 & 55.34 & 30.77 & 0.00 & 65.56 & 62.76 \\
SEA-LION v4.8 (Nemotron 30B)
    & 0.00 & 65.38 & 58.29 & 45.42 & 3.27 & 48.98 & 50.31 \\
\addlinespace
\multicolumn{8}{l}{\textit{Models $>$ 35B}} \\
Nemotron 3 Super
    & 0.00 & 53.07 & 48.82 & 41.47 & 18.81 & 50.18 & 35.49 \\
SEA-LION v4.8 (Nemotron 120B)
    & 37.68 & 71.72 & 67.85 & 54.88 & 47.47 & 61.99 & 50.84 \\

\midrule
\multicolumn{8}{l}{\textbf{Instruction Following}} \\
\multicolumn{8}{l}{\textit{Models $\leq$ 35B}} \\
Nemotron 3 Nano
    & 15.47 & 69.58 & 88.11 & 75.15 & 61.20 & 82.69 & 88.01 \\
Nemotron 3.5 Lightning
    & 16.40 & 50.87 & 89.17 & 59.30 & 31.15 & 78.56 & 80.78 \\
SEA-LION v4.8 (Nemotron 30B)
    & 53.13 & 74.42 & 80.27 & 82.06 & 70.01 & 79.96 & 85.92 \\
\addlinespace
\multicolumn{8}{l}{\textit{Models $>$ 35B}} \\
Nemotron 3 Super
    & 17.99 & 80.23 & 75.70 & 71.33 & 39.51 & 68.58 & 75.68 \\
SEA-LION v4.8 (Nemotron 120B)
    & 61.65 & 77.04 & 84.89 & 74.55 & 77.11 & 80.66 & 78.37 \\

\midrule
\multicolumn{8}{l}{\textbf{Knowledge}} \\
\multicolumn{8}{l}{\textit{Models $\leq$ 35B}} \\
Nemotron 3 Nano
    & 0.00 & 61.23 & 71.40 & 61.58 & -- & 57.71 & 63.33 \\
Nemotron 3.5 Lightning
    & 0.00 & 66.63 & 52.95 & 71.22 & -- & 61.19 & 50.82 \\
SEA-LION v4.8 (Nemotron 30B)
    & 0.68 & 68.25 & 76.11 & 53.28 & -- & 53.83 & 55.50 \\
\addlinespace
\multicolumn{8}{l}{\textit{Models $>$ 35B}} \\
Nemotron 3 Super
    & 0.00 & 49.90 & 75.70 & 76.10 & -- & 61.77 & 72.44 \\
SEA-LION v4.8 (Nemotron 120B)
    & 34.60 & 79.20 & 82.57 & 80.14 & -- & 62.80 & 74.65 \\

\midrule
\multicolumn{8}{l}{\textbf{Multi-turn}} \\
\multicolumn{8}{l}{\textit{Models $\leq$ 35B}} \\
Nemotron 3 Nano
    & -- & 71.51 & 79.37 & 77.73 & 44.84 & 79.77 & 77.05 \\
Nemotron 3.5 Lightning
    & -- & 77.62 & 75.96 & 82.80 & 50.16 & 81.41 & 83.20 \\
SEA-LION v4.8 (Nemotron 30B)
    & -- & 76.75 & 84.54 & 75.93 & 66.19 & 72.18 & 76.67 \\
\addlinespace
\multicolumn{8}{l}{\textit{Models $>$ 35B}} \\
Nemotron 3 Super
    & -- & 74.47 & 82.32 & 83.37 & 60.85 & 84.09 & 83.06 \\
SEA-LION v4.8 (Nemotron 120B)
    & -- & 79.22 & 82.70 & 81.37 & 71.68 & 82.60 & 83.06 \\

\bottomrule
\end{tabular}%
}
\caption{Performance comparison between Nemotron models and SEA-LION v4.8 across Southeast Asian languages: Cultural, Instruction Following, Knowledge, and Multi-turn.}
\vspace{-3mm}
\label{tab:sealion_v48_results_part1}
\end{table*}

\noindent
\textbf{Results.}
As shown in Tables~\ref{tab:sealion_v48_results_part1} and~\ref{tab:sealion_v48_results_part2}, \textsc{SEA-LION-v4.8} provides broad capability improvements over the Nemotron reference models, although the gains vary by task, language, and model scale.
For the 30B-A3B model, we compare against both Nemotron 3 Nano and Nemotron 3.5 Lightning, while the 120B-A12B model is compared against Nemotron 3 Super.
Overall, the 30B-A3B model shows particularly broad gains in {NLG} and {Instruction Following}, together with strong improvements in several NLR and NLU settings.
The 120B-A12B model exhibits more consistent gains across capabilities, particularly in {Cultural}, {Instruction Following}, {Knowledge}, {NLR}, and {NLU}.

For \textbf{Instruction Following}, the 30B-A3B model shows particularly large gains in Burmese, Filipino, Malay, and Tamil relative to both 30B reference models.
The 120B-A12B model similarly improves six of the seven languages over Nemotron 3 Super, with particularly large gains in Burmese and Tamil.
For \textbf{NLR}, the 30B-A3B model improves all available languages over Nemotron 3.5 Lightning, while the comparison with Nemotron 3 Nano is more mixed.
In contrast, the 120B-A12B model improves all available languages over Nemotron 3 Super, including large gains in Burmese and Tamil.
For \textbf{NLU}, Tamil shows the clearest improvement at both scales: the 30B-A3B model reaches 75.26 compared with 19.38 for Nemotron 3 Nano and 23.48 for Nemotron 3.5 Lightning, while the 120B-A12B model improves from 17.89 to 80.04.
The 120B-A12B model also substantially improves Burmese, Indonesian, and Malay.
The larger model further shows broad gains in \textbf{Cultural} and \textbf{Knowledge}: Cultural performance improves across all seven languages, while Knowledge improves across every language for which scores are available.

Beyond these broad improvements, the results also reveal complementary strengths across models and capabilities.
For the 30B-A3B model, \textbf{Cultural} performance improves over both 30B reference models in Filipino, Indonesian, and Malay, while the reference models remain stronger in Thai and Vietnamese.
A similar pattern is observed for \textbf{Knowledge}, where SEA-LION-v4.8 achieves its strongest gains in Filipino and Indonesian, while performance in Malay and Thai remains competitive with the reference models.
For \textbf{NLG}, the 30B-A3B model demonstrates particularly broad improvements, outperforming Nemotron 3 Nano across all seven languages and Nemotron 3.5 Lightning in six of the seven languages.
The 120B-A12B model also maintains strong NLG performance, improving over Nemotron 3 Super in six languages.
For \textbf{Safety}, SEA-LION-v4.8 shows encouraging gains in several languages: the 30B-A3B model improves Filipino and Vietnamese over both 30B references and substantially improves Thai over Nemotron 3.5 Lightning, while the 120B-A12B model improves Burmese, Malay, Tamil, Thai, and Vietnamese.
Finally, \textbf{Multi-turn} evaluation highlights notable gains in selected languages.
The 30B-A3B model improves Tamil from 44.84 for Nemotron 3 Nano and 50.16 for Nemotron 3.5 Lightning to 66.19, while the 120B-A12B model improves Filipino, Indonesian, and Tamil over Nemotron 3 Super.
%

Taken together, these results demonstrate that SEA-LION-v4.8 provides broad capability gains beyond the original Nemotron models.
The 30B-A3B model shows particularly strong improvements in instruction following, generation, and selected reasoning and understanding tasks, while the 120B-A12B model exhibits broader and more consistent gains across languages and capabilities.
Overall, the results show that our continued pre-training and post-training pipeline effectively strengthens Southeast Asian capabilities while maintaining strong performance across a diverse set of tasks.

\begin{table*}[h!]
\centering
\small
\setlength{\tabcolsep}{5pt}
\scalebox{0.8}{
\begin{tabular}{lrrrrrrr}
\toprule
\textbf{Model} &
\textbf{MY} & \textbf{TL} & \textbf{ID} & \textbf{MS} &
\textbf{TA} & \textbf{TH} & \textbf{VI} \\
\midrule

\multicolumn{8}{l}{\textbf{NLG}} \\
\multicolumn{8}{l}{\textit{Models $\leq$ 35B}} \\
Nemotron 3 Nano
    & 7.17 & 43.99 & 79.37 & 79.37 & 29.92 & 54.26 & 49.72 \\
Nemotron 3.5 Lightning
    & 10.14 & 44.92 & 75.96 & 81.16 & 22.24 & 57.04 & 51.89 \\
SEA-LION v4.8 (Nemotron 30B)
    & 18.97 & 46.49 & 84.54 & 90.30 & 37.66 & 54.93 & 53.25 \\
\addlinespace
\multicolumn{8}{l}{\textit{Models $>$ 35B}} \\
Nemotron 3 Super
    & 14.63 & 54.69 & 82.32 & 85.86 & 34.83 & 54.25 & 48.94 \\
SEA-LION v4.8 (Nemotron 120B)
    & 15.66 & 49.95 & 82.70 & 86.21 & 37.72 & 56.11 & 52.08 \\

\midrule
\multicolumn{8}{l}{\textbf{NLR}} \\
\multicolumn{8}{l}{\textit{Models $\leq$ 35B}} \\
Nemotron 3 Nano
    & 0.00 & 54.55 & 78.75 & -- & 14.45 & 64.44 & 60.92 \\
Nemotron 3.5 Lightning
    & 0.00 & 60.36 & 76.06 & -- & 5.98 & 44.01 & 40.11 \\
SEA-LION v4.8 (Nemotron 30B)
    & 1.47 & 62.60 & 82.58 & -- & 8.77 & 59.50 & 61.11 \\
\addlinespace
\multicolumn{8}{l}{\textit{Models $>$ 35B}} \\
Nemotron 3 Super
    & 0.00 & 59.90 & 80.43 & -- & 2.75 & 63.52 & 61.21 \\
SEA-LION v4.8 (Nemotron 120B)
    & 16.18 & 70.65 & 84.12 & -- & 40.98 & 66.41 & 66.52 \\

\midrule
\multicolumn{8}{l}{\textbf{NLU}} \\
\multicolumn{8}{l}{\textit{Models $\leq$ 35B}} \\
Nemotron 3 Nano
    & 0.00 & 58.30 & 70.47 & 52.56 & 19.38 & 56.95 & 66.78 \\
Nemotron 3.5 Lightning
    & 0.00 & 68.64 & 72.97 & 57.31 & 23.48 & 51.07 & 36.31 \\
SEA-LION v4.8 (Nemotron 30B)
    & 0.00 & 62.81 & 66.45 & 64.40 & 75.26 & 56.31 & 64.01 \\
\addlinespace
\multicolumn{8}{l}{\textit{Models $>$ 35B}} \\
Nemotron 3 Super
    & 2.25 & 68.43 & 63.35 & 55.17 & 17.89 & 56.49 & 63.49 \\
SEA-LION v4.8 (Nemotron 120B)
    & 43.46 & 65.67 & 77.92 & 68.72 & 80.04 & 57.93 & 65.85 \\

\midrule
\multicolumn{8}{l}{\textbf{Safety}} \\
\multicolumn{8}{l}{\textit{Models $\leq$ 35B}} \\
Nemotron 3 Nano
    & 0.00 & 32.84 & 54.40 & 27.32 & 0.00 & 56.66 & 40.45 \\
Nemotron 3.5 Lightning
    & 0.00 & 36.16 & 55.02 & 28.78 & 0.00 & 14.05 & 36.61 \\
SEA-LION v4.8 (Nemotron 30B)
    & 0.00 & 37.87 & 46.07 & 24.58 & 0.00 & 45.81 & 48.80 \\
\addlinespace
\multicolumn{8}{l}{\textit{Models $>$ 35B}} \\
Nemotron 3 Super
    & 0.00 & 45.18 & 55.19 & 21.35 & 0.00 & 30.40 & 47.42 \\
SEA-LION v4.8 (Nemotron 120B)
    & 10.20 & 41.89 & 49.32 & 34.73 & 65.55 & 32.28 & 47.50 \\

\bottomrule
\end{tabular}%
}
\caption{Performance comparison between Nemotron models and SEA-LION v4.8 across Southeast Asian languages: NLG, NLR, NLU, and Safety.}
\vspace{-3mm}
\label{tab:sealion_v48_results_part2}
\end{table*}

\section{Conclusion}

We introduced \textsc{Nemotron-SEA-LION-v4.8}, a family of Southeast-Asia-focused models at the 30B-A3B and 120B-A12B scales.
Our training pipeline combines continued pre-training on Southeast Asian and capability-oriented data with post-training based on supervised fine-tuning and online on-policy distillation.
On SEA-HELM, both models achieve higher overall SEA scores than their Nemotron baselines, with the 120B-A12B model showing particularly broad gains across all seven evaluated Southeast Asian languages.
Capability-level analysis shows strong improvements in instruction following, natural language generation, reasoning, and language understanding, while the 120B-A12B model additionally demonstrates broad gains in cultural and knowledge-oriented capabilities.
These results demonstrate that targeted continued pre-training and online post-training can substantially strengthen Southeast Asian language capabilities without requiring a new foundation model to be trained from scratch.

\section{Limitations}

SEA-LION-v4.8 has several limitations.
First, we retain the original Nemotron tokenizer, which is considerably less efficient for several Southeast Asian scripts. In particular, Khmer, Lao, and Tamil require substantially more tokens than with previous SEA-LION tokenizers, potentially limiting the effectiveness of additional language-specific training data.
Second, the continued pre-training corpus does not explicitly include Filipino data. Improvements for Filipino rely primarily on cross-lingual transfer rather than direct continued pre-training.
Third, although SEA-LION-v4.8 improves overall SEA-HELM performance, the gains are not uniform across capabilities. 
Instruction following, reasoning, and language understanding show the clearest improvements, while cultural, generation, safety, knowledge, and multi-turn tasks exhibit more language-dependent trade-offs.
Finally, our evaluation currently covers seven Southeast Asian languages and therefore does not represent the full linguistic diversity of the region. 

\section{Contributor}

Adila Aulia, Ahmed Dabeer, Ahn Jeongmi, Antonyrex Sajeban, Chan Hok Teng Adwin, Cheng Zi Yi Nicholas, Choa Hsueh Mei Esther, Heng Jonathan, Jann Railey Estrada Montalan, Lee Chwan Ren, Leong Wai Yi, Leong Wei Qi, Liew Rachel, Limkonchotiwat Peerat, Muhammad Ridzuan Bin Mokhtar, Nagarajan Karthik, Ng Boon Cheong Raymond, Ngee Chia Tai, Ngui Jian Gang, Nguyen Thanh Ngan, Ong Tat-Wee David, Pereira Mark, Phang Shi Wei Benjamin, Poon Joseph, Rengarajan Hamsawardhini, Susanto Yosephine, Sutaveephamochanon Anocha, Tan Choon Meng, Tan Chor Phin Evelyn, Tan Le Min Sheryl, Tan Siao Wei Jessica, Tan Yixian, Tasawong Panuthep, Tee Jun Yun, Teng Kok Wai Walter, Teo Eng Sipp Leslie, Tjhi William, Tuchinda Pume, Wu Donghang, Yong Xianbin, Zhang Zhou.

\section{Acknowledgments}
This project is supported by the National Research Foundation, Singapore under its National Large Language Models Funding Initiative. Any opinions, findings and conclusions or recommendations expressed in this material are those of the author(s) and do not reflect the views of National Research Foundation, Singapore.

\newpage

\bibliographystyle{plainnat}
{\small
\bibliography{custom}
}

\end{document}